\documentclass[10pt,twocolumn,letterpaper]{article}

\usepackage{iccv}
\usepackage{times}
\usepackage{epsfig}
\usepackage{graphicx}
\usepackage{amsmath}
\usepackage{amssymb}
\usepackage{enumitem, tabularx, booktabs}
\usepackage{fixltx2e}
\newcolumntype{Y}{>{\centering\arraybackslash}X}

\usepackage[pagebackref=true,breaklinks=true,letterpaper=true,colorlinks,bookmarks=false]{hyperref}

\iccvfinalcopy 

\def\iccvPaperID{1439} 

\ificcvfinal\fi
\begin{document}

\title{Deep Optics for Monocular Depth Estimation and 3D Object Detection}

\author{Julie Chang\\
Stanford University\\
{\tt\small jchang10@stanford.edu}
\and
Gordon Wetzstein\\
Stanford Univsersity\\
{\tt\small gordon.wetzstein@stanford.edu}
}

\maketitle

\begin{abstract}
Depth estimation and 3D object detection are critical for scene understanding but remain challenging to perform with a single image due to the loss of 3D information during image capture. Recent models using deep neural networks have improved monocular depth estimation performance, but there is still difficulty in predicting absolute depth and generalizing outside a standard dataset. Here we introduce the paradigm of deep optics, \ie end-to-end design of optics and image processing, to the monocular depth estimation problem, using coded defocus blur as an additional depth cue to be decoded by a neural network. We evaluate several optical coding strategies along with an end-to-end optimization scheme for depth estimation on three datasets, including NYU Depth v2 and KITTI. We find an optimized freeform lens design yields the best results, but chromatic aberration from a singlet lens offers significantly improved performance as well. We build a physical prototype and validate that chromatic aberrations improve depth estimation on real-world results. In addition, we train object detection networks on the KITTI dataset and show that the lens optimized for depth estimation also results in improved 3D object detection performance.

\end{abstract}


\section{Introduction}
\label{sec:introduction}

Depth awareness is crucial for many 3D computer vision tasks, including semantic segmentation~\cite{ren2012rgb,silberman2012indoor,gupta2013perceptual}, 3D object detection~\cite{shrivastava2013building,lin2013holistic,gupta2014learning,song2014sliding,song2016deep}, 3D object classification~\cite{wu20153d,maturana2015voxnet,qi2016volumetric}, and scene layout estimation~\cite{zhang2013estimating}. The required depth information is usually obtained with specialized camera systems, for example using time-of-flight, structured illumination, pulsed LiDAR, or stereo camera technology. However, the need for custom sensors, high-power illumination, complex electronics, or bulky device form factors often makes it difficult or costly to employ these specialized devices in practice.

\begin{figure}
\includegraphics[width=\linewidth]{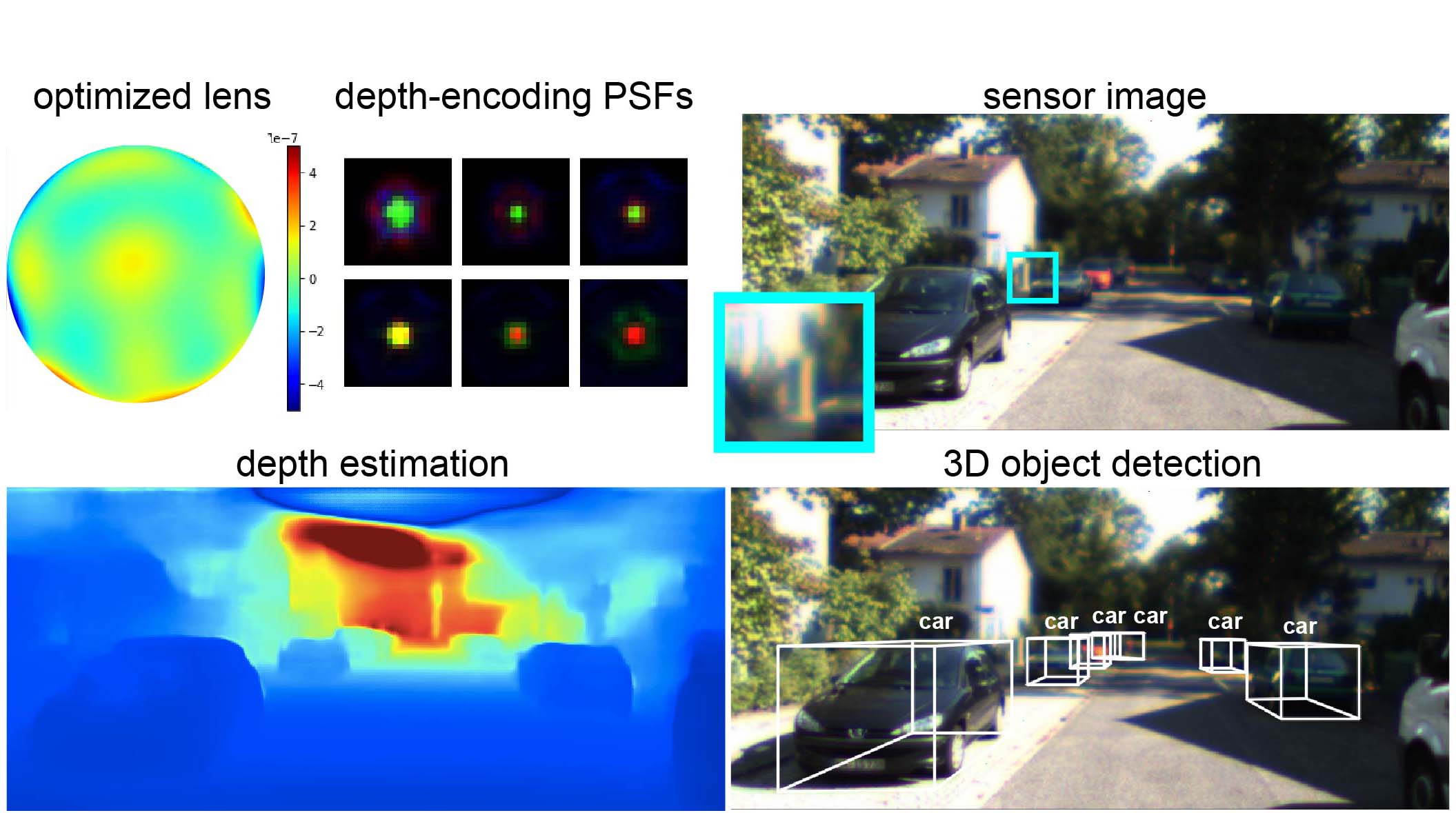}
\caption{We apply deep optics, \ie end-to-end design of optics and image processing, to build an optical-encoder, CNN-decoder system for improved monocular depth estimation and 3D object detection. }
\label{fig:teaser}
\end{figure}

Single-image depth estimation with conventional cameras has been an active area of research. Traditional approaches make use of pre-defined image features that are statistically correlated with depth, \eg shading, perspective distortions, occlusions, texture gradients, and haze \cite{horn1975obtaining,saxena2006learning,hoiem2007recovering,stella2008inferring,saxena2009make3d,ladicky2014pulling}. Recently, significant improvements have been achieved by replacing hand-crafted features with learned features via convolutional neural networks (CNNs) \cite{eigen2014depth,laina2016deeper,godard2017unsupervised,fu2018deep}. While these methods tend to perform decently within consistent datasets, they do not generalize well to scenes that were not part of the training set. In essence, the problem of estimating a depth map from pictorial cues alone is ill-posed. Optically encoding depth-dependent scene information has the potential to remove some of the ambiguities inherent in all-in-focus images, for example using (coded) defocus blur~\cite{pentland1987new,nayar1994illumination,levin2007image,veeraraghavan2007dappled,carvalho2018deep} or chromatic aberrations~\cite{trouve2013passive}. However, it is largely unclear how different optical coding strategies compare to one another and what the best strategy for a specific task may be. 

Inspired by recent work on deep optics \cite{chang2018hybrid,sitzmann2018end,haim2018depth}, we interpret the monocular depth estimation problem with coded defocus blur as an optical-encoder, electronic-decoder system that can be trained in an end-to-end manner. Although co-designing optics and image processing is a core idea in computational photography, only differentiable estimation algorithms, such as neural networks, allow for true end-to-end computational camera designs. Here, error backprograpagation in the training phase not only optimizes the network weights but also the physical lens parameters. With the proposed deep optics approach, we evaluate several variants of optical coding strategies for two important 3D scene understanding problems: monocular depth estimation and 3D object detection. 
 
In a series of experiments, we demonstrate that the deep optics approach optimizes the accuracy of depth estimation across several datasets. Consistent with previous work, we show that optical aberrations that are typically considered undesirable for image quality are highly beneficial for encoding depth cues. Our results corroborate that defocus blur provides useful information, but we additionally find that adding astigmatism and chromatic aberrations even further improves accuracy. By jointly optimizing a freeform lens, i.e. the spatially varying surface height of a lens, with the CNNs weights we achieve the best results. Surprisingly, we find that the accuracy of optimized lenses is only slightly better than standard defocus with chromatic aberrations. This insight motivates the use of simple cameras with only a single lens over complex lens systems when prioritizing depth estimation quality, which we validate with an experimental prototype. 

We also evaluate the benefits of deep optics for higher-level 3D scene understanding tasks. To this end, we train a PointNet \cite{qi2018frustum} 3D object detection network on the KITTI dataset. We find that, compared to all-in-focus monocular images, images captured through the optimized lenses also perform better in 3D object detection, a task which requires semantic understanding on top of depth estimation to predict 3D bounding boxes on object instances.

In sum, our experiments demonstrate that an optimized lens paired with a concurrently trained neural network can improve depth estimation without sacrificing higher-level image understanding. Specifically, we make the following contributions:
\begin{itemize}[leftmargin=*,noitemsep,topsep=0pt]
\item We build a differentiable optical image formation model that accounts for either fixed (defocus, astigmatism, chromatic aberration) or optimizable (freeform or annular) lens designs, which we integrate with a differentiable reconstruction algorithm, i.e. a CNN. 
\item We evaluate the joint optical-electronic model with the various lens settings on three datasets (Rectangles, NYU Depth-v2, KITTI). The optimized freeform phase mask yields the best results, with chromatic aberrations coming in a close second. 
\item We build a physical prototype and validate that captured images with chromatic aberrations achieve better depth estimation than their all-in-focus counterparts.
\item We train a 3D object detection network with the optimized lens and demonstrate that the benefits of improved depth estimation carry through to higher level 3D vision. 
\end{itemize}

Note that the objective of our work is not to develop the state-of-the-art network architecture for depth estimation, but to understand the relative benefits of deep optics over fixed lenses. Yet, our experiments show that deep optics achieves lower root-mean-square errors on depth estimation tasks with a very simple U-Net \cite{ronneberger2015u} compared to more complex networks taking all-in-focus images as input.


\section{Related Work}
\label{sec:related}

\paragraph{Deep Monocular Depth Estimation}

Humans are able to infer depth from a single image, provided enough contextual hints that allow the viewer to draw from past experiences. Deep monocular depth estimation algorithms aim at mimicking this capability by training neural networks to perform this task \cite{eigen2014depth,laina2016deeper,godard2017unsupervised,fu2018deep}. Using various network architectures, loss functions, and supervision techniques, monocular depth estimation can be fairly successful on consistent datasets such as KITTI~\cite{geiger2013vision} and NYU Depth~\cite{silberman2012indoor}. However, performance is highly dependent on the training dataset. To address this issue, several recent approaches have incorporated physical camera parameters into their image formation model, including focal length \cite{he2018learning} and defocus blur \cite{carvalho2018deep}, to implicitly encode 3D information into a 2D image. We build on these previous insights and perform a significantly more extensive study that evaluates several types of fixed lenses as well as fully optimizable camera lenses for monocular depth estimation and 3D object detection tasks.

\begin{figure*}
\includegraphics[width=\textwidth]{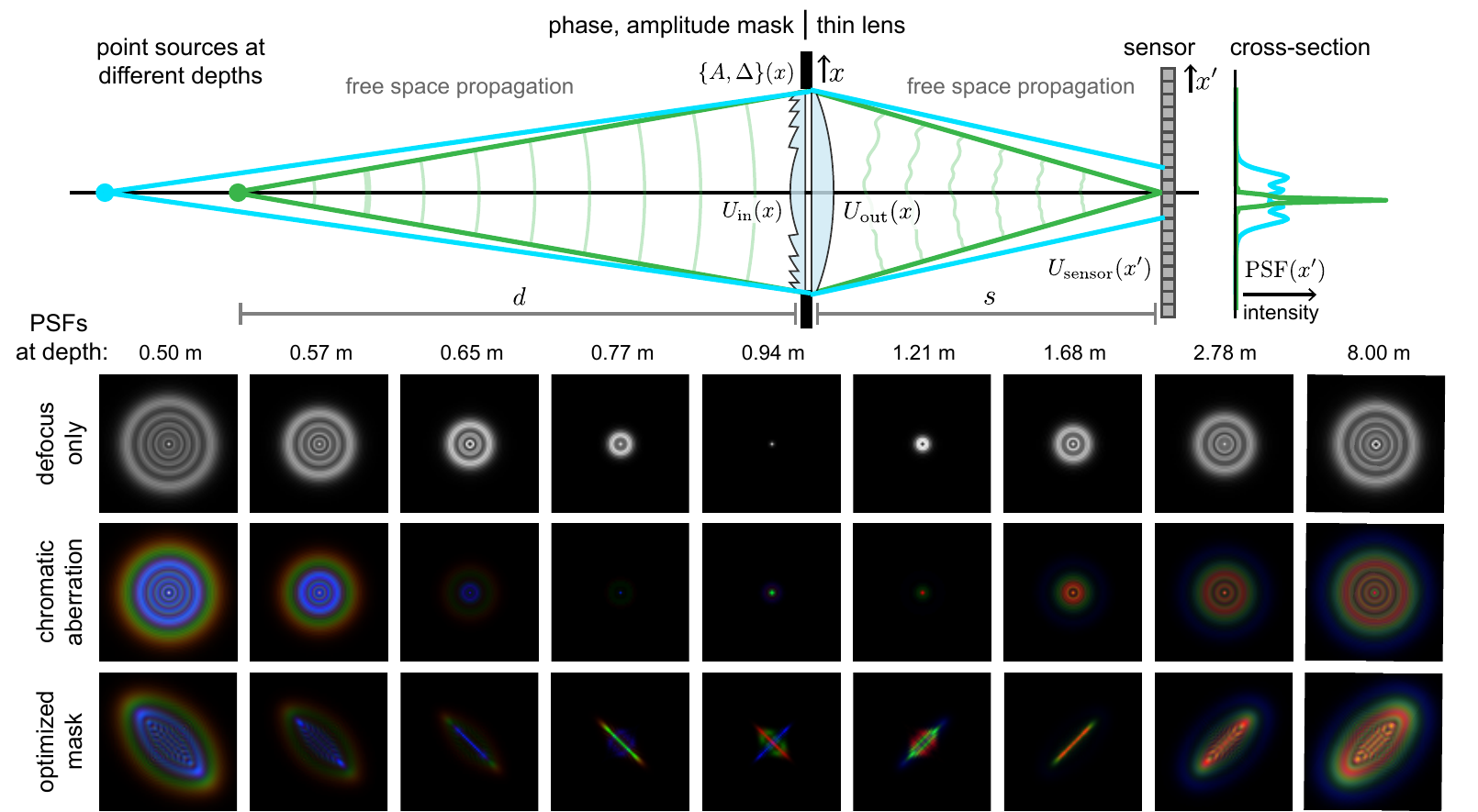}
\caption{\textbf{PSF simulation model.} (Top) Optical propagation model of point sources through a phase mask placed in front of a thin lens. PSFs are simulated by calculating intensity of the electric field at the sensor plane. (Bottom) Sample PSFs from thin lens defocus only, with chromatic aberrations, and using an optimized mask initialized with astigmatism. }
\label{fig:psfs}
\end{figure*}

\paragraph{Computational Photography for Depth Estimation}

Modifying camera parameters for improved depth estimation is a common approach in computational photography. For example, coding the amplitude \cite{levin2007image,veeraraghavan2007dappled,zhou2011coded} or phase \cite{levin20094d} of a camera aperture has been shown to improve depth reconstruction. Chromatic aberrations have also been shown to be useful for estimating the depth of a scene \cite{trouve2013passive}. Whereas conventional defocus blur is symmetric around the focal plane, i.e. there is one distance in front of the focal plane that has the same PSF as another distance behind the focal plane, defocus blur with chromatic aberrations is unambiguous. In all these approaches, depth information is encoded into the image in a way that makes it easier for an algorithm to succeed at a certain task, such as depth estimation. In this paper, we combine related optical coding techniques with more contemporary deep-learning methods. The primary benefit of a deep learning approach over previous work is that these allow a loss function applied to a high-level vision task, e.g. object detection, to directly influence physical camera parameters in a principled manner, such as the lens surface.
 

\paragraph{Deep Optics}

Deep learning can be used for jointly training camera optics and CNN-based estimation methods. This approach was recently demonstrated for applications in extended depth of field and superresolution imaging~\cite{sitzmann2018end}, image classification~\cite{chang2018hybrid}, and multicolor localization microscopy~\cite{michaeli2018multicolor}. For example, Hershko \etal~\cite{michaeli2018multicolor} proposed to learn a custom diffractive phase mask that produced highly wavelength-dependent point spread functions (PSFs), allowing for color recovery from a grayscale camera. In our applications, an optical lens model also creates depth-dependent PSFs with chromatic aberrations. However, our deep camera is designed for computer vision applications rather than microscopy. The work closest to ours is that of Haim \etal~\cite{haim2018depth}, who designed a diffractive phase mask consisting of concentric rings to induce chromatic aberrations that could serve as depth cues \cite{haim2018depth}. The training process optimized the ring radii and phase shifts within two or three annular rings but did not allow for deviation from this simple parametric lens model. In our experiments, we systematically evaluate the comparative performances of non-optimized aberrated lenses as well as fully optimizable freeform lenses. Unlike previous work, we explore applications in depth estimation and also 3D object detection.

\section{Differentiable Image Formation Model}
\label{sec:imageformation}
To optimize optical lens elements that best encode depth-dependent scene information, we model light transport in the camera using wave optics. This is not only physically accurate but also allows for both refractive and diffractive optical elements to be optimized. Due to the fact that the light in a natural scene is incoherent, we only rely on a coherent light transport model to simulate the depth- and wavelenth-dependent point spread function (PSF) of the system, which we then use to simulate sensor images. 

\subsection{Modeling Conventional Cameras}

We begin by building a camera model consisting of a single convex thin lens with focal length $f$ at a distance $s$ from the sensor (see Fig.~\ref{fig:psfs}). The relationship between the in-focus distance and the sensor distance is given by the thin-lens equation:
\begin{equation}
\frac{1}{f} = \frac{1}{d} + \frac{1}{s}
\end{equation}
Hence an object at a distance $d$ in front of the lens appears in focus at a distance $s$ behind the lens. 


When imaging a real-world scene, there are likely to be objects at multiple depths that are imaged with different PSFs. To simulate the PSF at a depth $z$, we consider a point emitter of wavelength $\lambda$ centered on the optical axis located a distance $z$ away from the center of the thin lens. Our general approach is to propagate the wave of light through the optical system to the sensor. To begin, we first propagate the light emitted by the point, represented as a spherical wave, to the lens. The complex-valued electric field immediately before the lens is given by:
\begin{equation}
U_\text{in}(x, y) = \exp(i k \sqrt{x^2 + y^2 + z^2})
\end{equation}
where $k = 2\pi / \lambda$ is the wavenumber. 

The next step is to propagate this wave field through the lens by multiplying the input by a phase delay, $t(x,y)$, induced at each location on the lens. Such a phase shift of a wave is physically produced by light slowing down as it propagates through the denser material of the optical element. The thickness profile, $\Delta(x,y)$, of a convex thin lens with index of refraction $n(\lambda)$ in a paraxial regime \cite{goodman} is 
\begin{equation}\label{eq:lens}
\Delta(x,y) = \Delta_0 - \frac{x^2+y^2}{2f(n(\lambda)-1)}
\end{equation}
where $\Delta_0$ is the center thickness. Note that the refractive index is wavelength-dependent, which is necessary to model chromatic aberrations correctly. Converting thickness to the corresponding phase shift, $\phi = k(n-1)\Delta$, and neglecting the constant phase offset from $\Delta_0$, the phase transformation is
\begin{equation}
t(x,y) = e^{i\phi(x,y)} = \exp\left[-i\frac{k}{2f}(x^2+y^2)\right]
\end{equation}
Additionally, since a lens has some finite aperture size, we insert an amplitude function $A(x,y)$ that blocks all light in regions outside the open aperture. To find the electric field immediately after the lens, we multiply the amplitude and phase modulation of the lens with the input electric field:
\begin{equation}
U_\text{out}(x,y) = A(x,y)\ t(x,y)\ U_\text{in}(x,y)
\end{equation}
Finally, the field propagates a distance $s$ to the sensor with the exact transfer function \cite{goodman}:
\begin{equation}
 H_\text{s}(f_x, f_y) = \exp\left[i k s\sqrt{1 - (\lambda f_x)^2 - (\lambda f_y)^2}\right]
\end{equation}
where $(f_x, f_y)$ are spatial frequencies. This transfer function is applied in the Fourier domain as:
\begin{equation}
U_\text{sensor}(x',y') = \mathcal{F}^{-1}\big\{
\mathcal{F}\left\{ U_\text{out}(x,y) \right\} \cdot H_\text{s}(f_x, f_y) 
\big\}
\end{equation}
where $\mathcal{F}$ denotes the 2D Fourier transform. Since the sensor measures light intensity, we take the magnitude-squared to find the final PSF:
\begin{equation}
\text{PSF}_{\lambda, z} (x',y') = |U_\text{sensor}(x',y')|^2
\end{equation}
By following this sequence of forward calculations, we can generate a 2D PSF for each depth and wavelength of interest. Since the lens was initially positioned to focus at a distance $d$, we can expect the PSF for $z=d$ to have the sharpest focus and to spread out away from this focal plane.

\begin{figure*}[t]
\includegraphics[width=\textwidth]{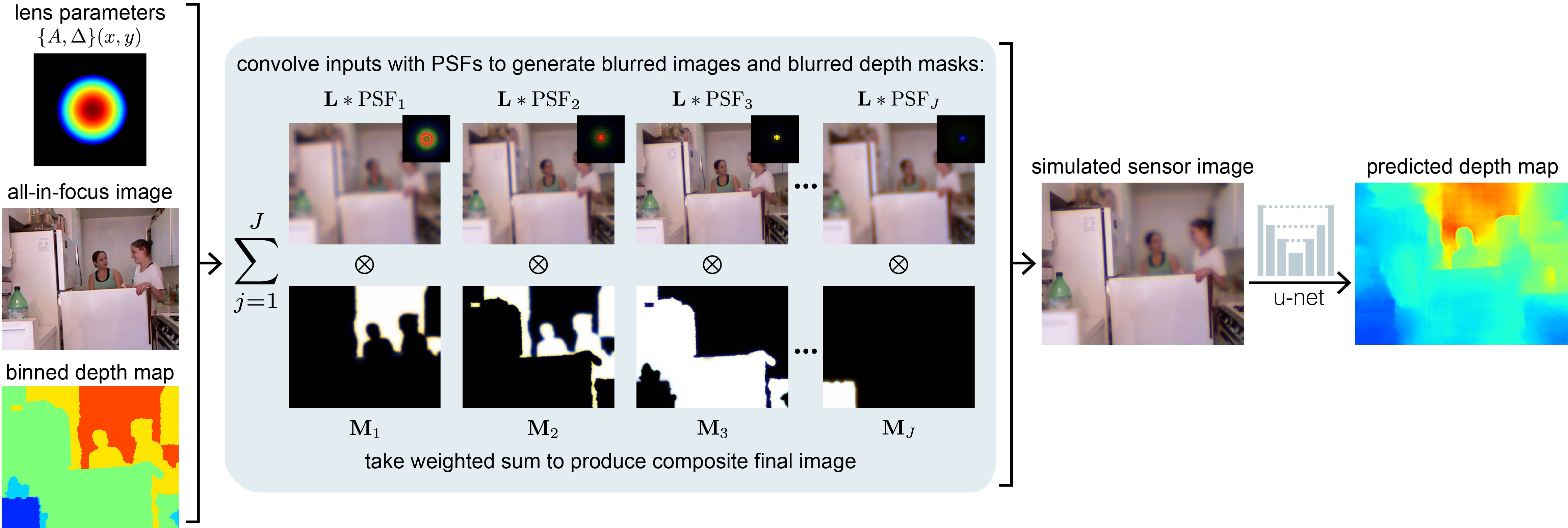}
\caption{\textbf{Depth-dependent image formation.} Given a set of lens parameters, an all-in-focus image, and its binned depth map, the image formation model generates the appropriate PSFs and applies depth-dependent convolution to simulate the corresponding sensor image, which is then passed into a U-Net for depth estimation.}
\label{fig:conv}
\end{figure*} 

%
%

\subsection{Modeling Freeform Lenses}

Several variables such as focal length, focus distance, and aperture size are modeled by the above formulation. For maximum degrees of freedom to shape the PSF, we can also treat the optical element as a freeform lens by assuming that is has an additional arbitrary thickness profile $\Delta_\text{ff}(x,y)$. The corresponding phase delay is 
%
%
%
\begin{equation}
t_\text{ff}(x,y) = \exp\left[jk(n_\text{ff}(\lambda)-1)\Delta_\text{ff}(x,y) \right]
\end{equation}
where $n_\text{ff}(\lambda)$ is the wavelength-dependent index of refraction of the lens material. We parametrize $\Delta_\text{ff}$ with the Zernike basis (indices 1-36, \cite{noll1976zernike}), which leads to smoother surfaces. The intensity PSF of a freeform lens is then
\begin{equation}
\text{PSF}_{\lambda, z} (x,y; \lambda) = |\mathcal{F}^{-1}\{\mathcal{F}\{A\cdot t_\text{lens} \cdot  t_\text{ff} \cdot U_\text{in}\} \cdot H_\text{s} \}|^2 (x,y)
\end{equation}


\subsection{Depth-Dependent Image Formation}

We can use these simulated PSFs to approximate a captured image of a 3D scene on an RGB sensor. 
To this end, we use a layered representation that models the scene as a set of planar surfaces at a discrete number of depth planes \cite{hasinoff2007layer}. This allows for precomputation of a fixed number of PSFs corresponding to each depth plane. We make a few modifications here to suit our datasets consisting of pairs of all-in-focus RGB images and their discretized depth maps. 
For an all-in-focus image $\textbf{L}$, a set of $j=1 \ldots J$ discrete depth layers, and occlusion masks $\{\textbf{M}_j\}$, we calculate our final image by:
\begin{equation}
\textbf{I}_{\lambda} = \sum_{j=1}^J (\textbf{L}_{\lambda} * \text{PSF}_{\lambda,j}) \cdot \textbf{M}_j 
\end{equation}
where $*$ denotes 2D convolution for each color channel centered on $\lambda$. The occlusion masks $\{\textbf{M}_j\}$ represent the individual layers of the quantized depth map. To ensure smooth transitions between the masks of a scene, we additionally blur each of the quantized layers and re-normalize them, such that $\sum_j \textbf{M}_j = 1$ at each pixel.

%
%
%
%

\section{Depth Estimation} 
\label{sec:depth}
In this section, we detail our experiments for deep optics for monocular depth estimation with encoded blur.

\subsection{Network and Training}

For depth estimation, we connect our differentiable image formation model to a U-Net~\cite{ronneberger2015u} that takes as input either the simulated sensor images or the original all-in-focus dataset images. The network consists of 5 downsampling layers (\{Conv-BN-ReLU\}$\times$2$\rightarrow$MaxPool2$\times$2) followed by 5 upsampling layers with skip connections (Conv$^T$+Concat$\rightarrow$\{Conv-BN-ReLU\}$\times$2). The output is the predicted depth map, at the same resolution as the input image. We use the standard ADAM optimizer with a mean-square-error (MSE) loss on the logarithmic depth. We train the models for 40,000 iterations at a learning rate of .001 and batch size of 3. We additionally decay the learning rate to 1e-4 for the Rectangles dataset.

We evaluate on (1) a custom Rectangles dataset, which consists of white rectangles against a black background places at random depths (see Supplement), (2) the NYU Depth v2 dataset with standard splits, and (3) a subset of the KITTI depth dataset (5500 train, 749 val) that overlaps with the object detection dataset for which we obtained dense ``ground truth" depth maps from Ma \etal \cite{mal2018sparse}. We train on full-size images. We calculate loss for NYU Depth on the standard crop size, and for KITTI only on the official sparse ground truth depth.

\begin{table*}[]
\begin{center}
\begin{tabularx}{.9\textwidth}{ l|| *{5}{Y|}Y }
\hline

          &\multicolumn{2}{c | }{\textbf{Rectangles}} &\multicolumn{2}{c|}{\textbf{NYU Depth v2}} & \multicolumn{2}{c}{\textbf{KITTI*}} \\ 
\textbf{Optical model}          & RMSE$_\text{lin}$ & RMSE$_\text{log}$ & RMSE$_\text{lin}$ & RMSE$_\text{log10}$ & RMSE$_\text{lin}$ & RMSE$_\text{log}$\\ 
\hline
\hline
All-in-focus			& 0.4626	& 0.3588	& 0.9556	& 0.1452	& 2.9100	& 0.1083	\\ \hline
Defocus, achromatic     	& 0.2268	& 0.1805	& 0.4814	& 0.0620	& 2.5400	& 0.0776	  \\ \hline
Astigmatism, achromatic & 0.1348	& 0.0771	& 0.4561	& 0.0559	& 2.3634	& 0.0752	  \\ \hline
Chromatic aberration 	&  \it{0.0984}	&  \it{0.0563}	& \it{0.4496}	& \it{0.0556}	&  \it{2.2566}	& \it{0.0702}	    \\ \hline
Optimized, annular 		& 0.1687	& 0.1260	& 0.4817	& 0.0623	& 2.7998	& 0.0892  \\ \hline 
Optimized, freeform 	& \bf{0.0902}	& \bf{0.0523}	& \bf{0.4325}	& \bf{0.0520}	& \bf{1.9288} & \bf{0.0621} \\ \hline
\end{tabularx}
\vspace{10pt}
\caption{Depth estimation error with different optical models for various datasets. RMSEs are reported for linear and log (base $e$ or 10) scaling of depth (m or log(m)). Lowest errors are bolded, and second-lowest are italicized. The KITTI* dataset is our KITTI dataset subset.}

\label{table:summary}
\end{center}
\end{table*}

For the Rectangles and NYU Depth datasets, we initialize the phase mask as an f/8, 50 mm focal length lens, focused to 1 m. For the KITTI dataset, we initialize an f/8, 80 mm focal length lens, focused to 7.6 m.  When the lens is being optimized, we also initialize the U-Net with the optimized weights for the fixed lens, and each training step adjusts the parameters of the lens (Zernike coefficients for freeform, ring heights for annular) and the U-Net.  We use 12 depth bins in our simulations, spaced linearly in inverse depth. When optimizing a freeform lens for the KITTI dataset, we reduce this to 6 intervals due to GPU memory constraints and train for 30,000 iterations; then we freeze the lens and increase back to 12 intervals to fine-tune the U-Net for an additional 30,000 iterations.


\subsection{Analysis and Evaluation} 

Table \ref{table:summary} shows a summary of results for all datasets. Examples of simulated sensor images and predicted depth maps from NYU Depth and KITTI are shown in Fig. \ref{fig:depth} (see Supplement for Rectangles). 

We observe common trends across all datasets. When using the all-in-focus images, errors are highest. This is most intuitive to understand with the Rectangles dataset. If there is a randomly-sized white rectangle floating in space that is always in focus, there are no depth cues for the network to recognize, and the network predicts the mean depth for every rectangle. Depth from defocus-only improves performance, but there is still ambiguity due to symmetric blur along inverse depth in both directions from the focal plane. Astigmatism (see Supplement for details) helps resolve this ambiguity, and the inherent chromatic aberration  of a singlet lens further improves results. 

We optimize two freeform lenses for each dataset. The annular lens consist of three concentric layers of different heights, inspired by \cite{haim2018depth}. While these optimized lenses outperformed all-in-focus experiments, they did not yield higher accuracy than chromatic aberration from a fixed lens. In contrast, the optimized freeform lens showed the best results, demonstrating the ability of the end-to-end optimization to learn a new freeform lens that better encodes depth information.  For NYU Depth, we found that additionally initializing $\Delta_\text{ff}$ with astigmatism yielded the better results. 

Table \ref{table:nyu} additionally compares default metrics on the NYU Depth test set with reported results from previous works. These comparisons suggest that adding this optical encoder portion of the model can yield results on par with state-of-the-art methods with more heavyweight and carefully designed networks. 

\begin{table}[]
\setlength\tabcolsep{1.8pt}
\begin{tabularx}{\linewidth}{l || *{3}{Y} | *{3}{Y} } 
\hline
\textbf{Method}	& rel & log10 & rms & $\delta_1$ & $\delta_2$ & $\delta_3$ \\ \hline \hline
Laina \etal \cite{laina2016deeper}     & 0.127 & 0.055 &  0.573 & 0.811 & 0.953 & 0.988 \\ 
MS-CRF \cite{xu2017multi}     & 0.121 & 0.052 &  0.586 & 0.811 & 0.954 & 0.987 \\ 
DORN \cite{fu2018deep}    	& 0.115 & \textbf{0.051} & 0.509 & 0.828 & 0.965 & 0.992 \\
\hline
All-in-focus    & 0.293 & 0.145 & 0.956 & 0.493 & 0.803 & 0.936     \\ 
Defocus     	& 0.108 & 0.062 & 0.481 & 0.893 & 0.981 & 0.996 \\ 
Astigmatism	& 0.095 & 0.056 & 0.456 & 0.916 & 0.986 & 0.998 \\ 
Chromatic 	& 0.095 & 0.056 & 0.450 & 0.916 & 0.987 & 0.998 \\ 
Freeform 		& \textbf{0.087} & 0.052 & \textbf{0.433} & \textbf{0.930} & \textbf{0.990} & \textbf{0.999} \\ \hline 
\end{tabularx}
\vspace{10pt}
\caption{Comparative performance on NYU Depth v2 test set, as calculated in \cite{eigen2014depth}. Units are in meters or log10(m). Thresholds are denoted $\delta_i: \delta > 1.25^i$. Lowest errors and highest $\delta$s are bolded.}
\label{table:nyu}
\end{table}

\begin{figure*}[ht]
\begin{center}
\includegraphics[width=.95\textwidth]{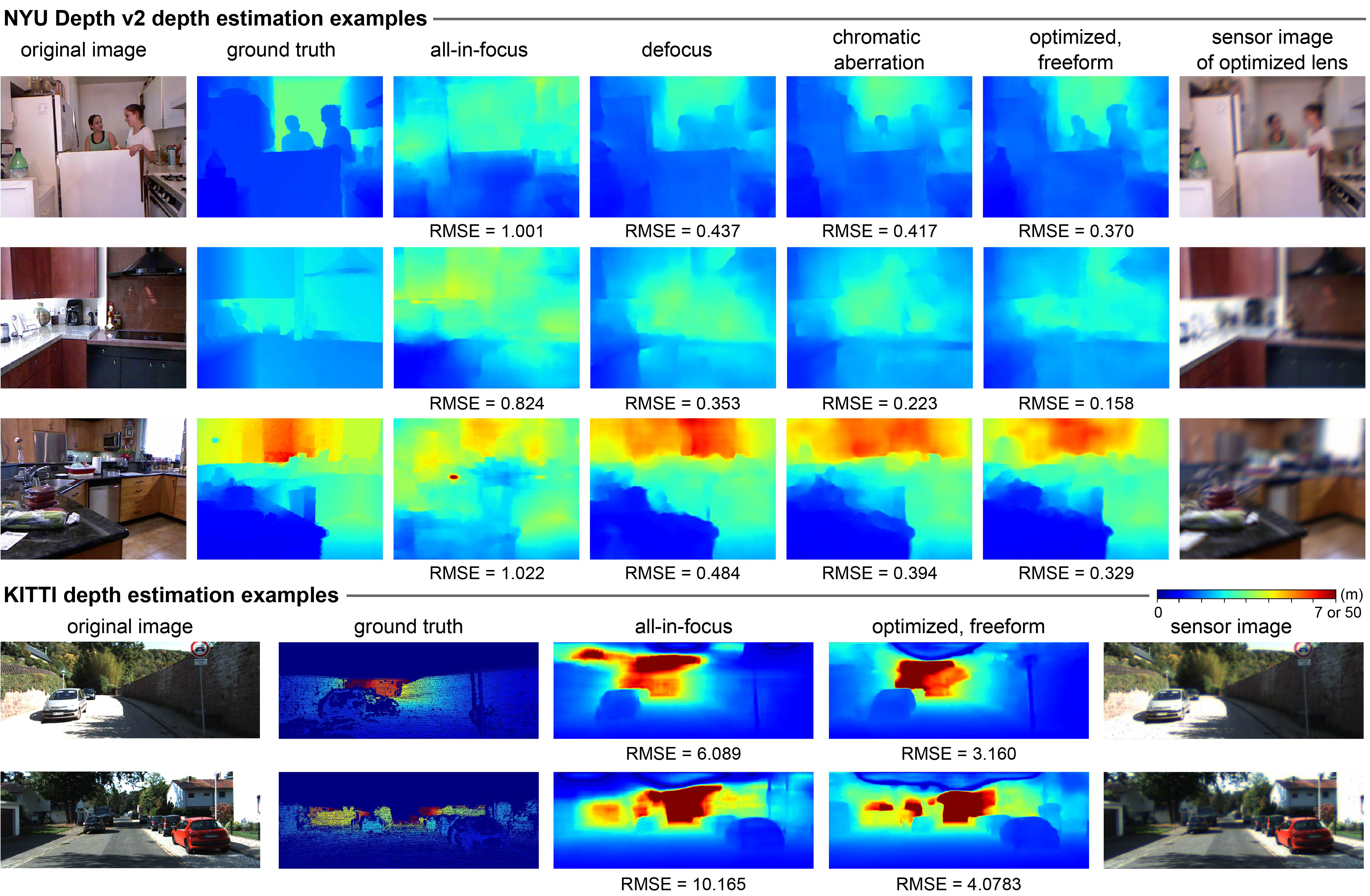}
\end{center}
\vspace{-6pt}
\caption{\textbf{Depth estimation.} (Top) Examples with RMSE (m) from the NYU Depth v2 dataset with all-in-focus, defocus, chromatic aberration, and optimized models. The simulated sensor image from the optimized system is also shown. (Bottom) Examples with RMSE (m) from the KITTI dataset (cropped to fit) with all-in-focus and optimized models; the sensor image from the optimized model is also shown. All depth maps use the same colormap, but the maximum value is 7 m for NYU Depth and 50 m for KITTI.}
\label{fig:depth}
\end{figure*}


\subsection{Experimental Results}

\begin{figure*}[ht]
\includegraphics[width=\textwidth]{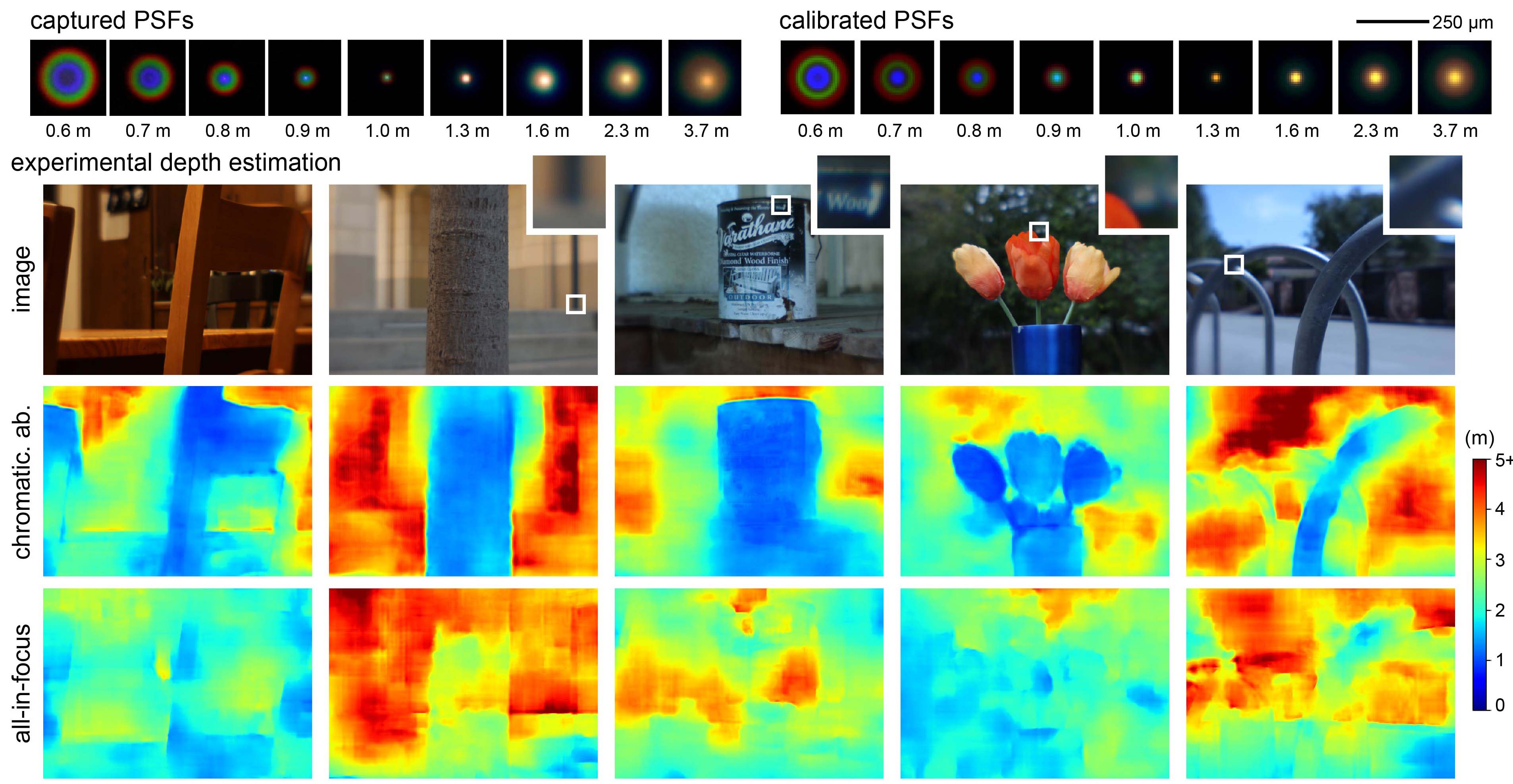}
\caption{\textbf{Real-world capture and depth estimation.} (Top) Captured and calibrated depth-dependent PSFs, displayed at the same scale. (Bottom) Examples of images captured using our prototype with a zoomed region inset, depth estimation with chromatic aberration, and depth estimation from the corresponding all-in-focus image (not shown). Depth map colorscale is the same for all depth maps. }
\label{fig:captured}
\end{figure*} 

We build a prototype for monocular depth estimation using chromatic aberration on real-world scenes. Our camera consisted of a Canon EOS Rebel T5 camera and a biconvex singlet lens ($f$ = 35mm, Thorlabs) with a circular aperture (D = 0.8 mm). We captured a series of images of a point white light source to calibrate the modeled PSFs with the captured PSFs, primarily by adjusting a spherical aberration parameter. We retrain a depth estimation network for the calibrated PSFs with the NYU Depth dataset, including a downsampling factor of four due to the smaller image size of dataset compared to the camera sensor. For this network, after convolution in linear intensity, we apply sRGB conversion to produce the simulated sensor image, which allows us to directly input captured sRGB camera images during evaluation. 

We capture images in a variety of settings with the prototype as described along with an all-in-focus pair obtained by adding a 1 mm pinhole in front of the lens (see Supplement for images). We use our retrained depth estimation network to predict a depth map from the blurry images, and we use the all-in-focus network to predict the corresponding depth map from the all-in-focus images. Fig. \ref{fig:captured} shows a few examples; more are included in the supplement. Depth estimation with the optical model performs significantly better on the captured images, as physical depth information is actually encoded into the image, allowing the network to rely not just on dataset priors for prediction. A limitation of our prototype was its smaller field of view, due to camera vignetting and the spatially varying nature of the real PSF, which prevented capture of full indoor room scenes. This could be improved by adding another lens to correct for other aberrations \cite{cossairt2010spectral} or by including these variations in the image formation model \cite{heide2013high}.

\section{3D Object Detection}
\label{sec:detection}

\begin{table}
\setlength\tabcolsep{1.8pt}
\begin{tabularx}{\linewidth}{l || Y|Y } 
\hline
\textbf{Object detection metric}	& \textbf{All-in-focus} & \textbf{Optimized} \\\hline \hline
2D mAP & 78.01 & \textbf{78.96} \\ 
2D AP, Car & \textbf{95.50} & 95.15\\ 
2D AP, Pedestrian & 80.06 & \textbf{80.22}\\ 
2D AP, Cyclist & \textbf{89.77} & 88.11 \\ \hline
3D AP, Ped., Easy & 9.74 & \textbf{13.86} \\
3D AP, Ped., Moderate & 7.10 & \textbf{11.74} \\
3D AP, Ped., Hard & 6.21 & \textbf{11.90} \\
3D AP, Cyc., Easy & 2.27 & \textbf{7.18} \\
3D AP, Cyc., Moderate & 2.36 & \textbf{4.89} \\
3D AP, Cyc., Hard & 1.98 & \textbf{4.95} \\ \hline
\end{tabularx}
\vspace{10pt}
\caption{Object detection performance measured by 2D AP \% (IoU = 0.5) and 3D AP \% (IoU = 0.5) on our validation split of the KITTI object detection dataset using the all-in-focus and optimized mask models. Higher values are bolded.}
\label{table:objval}
\end{table}

\begin{table*}[]
\begin{center}
\begin{tabular}{l | c || c | c | c | c | c | c }
\hline
 & &  \multicolumn{3 }{c | }{\textbf{3D object localization}}  & \multicolumn{3}{c}{\textbf{3D object detection}} \\ 
\textbf{Method} & \textbf{Input}   & Easy & Moderate & Hard & Easy & Moderate & Hard \\ 
\hline
\hline
Mono3D \cite{chen2016monocular}  &  RGB  &  5.22     & 5.19      & 4.13  & 2.53    & 2.31 & 2.31  \\ 
MF3D \cite{xu2018multi}     &  RGB    &   22.03	& 13.63	& 11.6	& 10.53	& 5.69 & 5.39  \\
MonoGRNet \cite{qin2018monogrnet}     &  RGB              &  -    & -       & -           & 13.88	& 10.19 & 	7.62 \\ \hline
VoxelNet \cite{zhou2018voxelnet} & RGB+LIDAR &   89.6	& 84.81	& 78.57	& 81.97 & 65.46&  62.85 \\ 
FPointNet \cite{qi2018frustum} & RGB+LIDAR              &   88.16 & 	84.02 &  76.44 &  83.76 &  70.92 & 63.65  \\ \hline
(Ours) All-in-focus (val)  &  RGB & 26.71  & 19.87 & 19.11 &  16.86 & 13.82 & 13.26 \\ 
(Ours) Optimized, freeform (val) & RGB & 37.51 & 25.83 & 21.05 & 25.20 & 17.07 & 13.43 \\ \hline
\end{tabular}
\vspace{10pt}
\caption{3D object localization AP \% (bird's eye view) and 3D object detection AP \% (IoU$ = 0.7$) for the car class. The listed numbers from literature are reported on the official test set; results from our methods are reported on our validation split.}
\label{table:objtest}
\end{center}
\end{table*}

To assess whether an optical system optimized for improved depth estimation is beneficial for higher-level 3D scene understanding as well, we evaluate 3D object detection performance on the KITTI dataset using the same optical system. 3D object detection requires recognizing instances of different objects as well as regressing an oriented 3D bounding box around each object instance. Depth information, whether implicitly contained in an image or explicitly provided from a depth sensor, is critical for this task, as is evidenced in the large gap in performance between the RGB and RGB+LIDAR methods shown in Table \ref{table:objtest}. 

We train a 3D object detection network specific to the freeform lens optimized for KITTI depth estimation. In particular, we use a Frustrum PointNet v1 (FPointNet, \cite{qi2018frustum}), which was demonstrated to work with both sparse LIDAR point clouds and dense depth images. FPointNet first uses 2D bounding box predictions on the RGB image to generate frustrum proposals that bound a 3D search space; then 3D segmentation and box estimation occur on the 3D point cloud contained within each frustrum. In our modified network, we substitute the ground truth LIDAR point clouds with our estimated depth maps projected into a 3D point cloud. As in the original method, ground truth 2D boxes augmented with random translation and scaling are used during training, but estimated 2D bounding boxes from a separately trained 2D object detection network (Faster R-CNN, \cite{NIPS2015_5638}) are used during validation. Since we require accurate dense ground truth depth maps to generate our simulated sensor images, we report results for our validation split, for which we did obtain a reliable dense depth map. For comparison, we train the same networks with all-in-focus images and their estimated depth maps. More details on our implementation of these networks and assessment on the test set is included in the Supplement. 

Results of our object detection experiments are shown in Tables \ref{table:objval} and \ref{table:objtest}. Average precision (AP) values are computed by the standard PASCAL protocol, as described in the KITTI development kit. 2D object detection performance is similar between the all-in-focus and optimized systems, which implies that even though the sensor images from the optimized optical element appear blurrier than the all-in-focus images, the networks are able to extract comparable information from the two sets of images. More notably, 3D object detection improves with the optimized optical system, indicating that the FPointNet benefits from the improved depth maps enabled with the optimized lens.

\section{Discussion}
\label{sec:discussion}
Throughout our experiments, we demonstrate that a joint optical-encoder, electronic-decoder model outperforms the corresponding optics-agnostic model using all-in-focus images. We build a differentiable optical image formation layer that we join with a depth estimation network to allow for end-to-end optimization from camera lens to network weights. The fully optimized system yields the most accurate depth estimation results, but we find that native chromatic aberrations can also encode valuable depth information. Additionally, to verify that improved depth encoding does not need to sacrifice other important visual content, we show that the lens optimized for depth estimation maintains 2D object detection performance while further improving 3D object detection from a single image.

As mentioned, our conclusions are primarily drawn from the relative performance between our results. We do not claim to conclusively surpass existing methods, as we use the ground truth or pseudo-truth depth map in simulating our sensor images, and we are limited to an approximate, discretized, layer-based image formation model. There may be simulation inaccuracies that are not straightforward to disentangle unless the entire dataset was recaptured through the different lenses. Nonetheless, our real-world experimental results are promising in supporting the advantage of optical depth encoding, though more extensive experiments, especially with a larger field-of-view, would be valuable. We are interested in future work to see how an optical layer can further improve leading methods, whether for monocular depth estimation \cite{laina2016deeper,xu2017multi,fu2018deep} or other visual tasks.

More broadly, our results consistently support the idea that incorporating the camera as an optimizable part of the network offers significant benefits over considering the image processing completely separately from image capture. We have only considered the camera as a single static optical layer in this paper, but there may be potential in more complex designs as research in both optical computing and computer vision continues to advance.


{\small
\bibliographystyle{ieee}
\bibliography{bibliography}
}

\end{document}